\documentclass[11pt]{article}

\usepackage[T1]{fontenc}
\usepackage[utf8]{inputenc}
\usepackage{lmodern}
\usepackage[margin=1in]{geometry}
\usepackage[numbers,sort&compress]{natbib}
\usepackage{amsmath,amssymb}
\usepackage{graphicx}
\usepackage{booktabs}
\usepackage{array}
\usepackage{microtype}
\usepackage{caption}
\usepackage[hidelinks]{hyperref}
\graphicspath{{figures/}}

\title{BioMed Harness: A Traceable, Self-Improving Agentic Platform for Liquid-Biopsy Biomarker Discovery and Patient Data Analytics}

\author{Sophia Y. Liu\textsuperscript{1,2}, Xiang Li\textsuperscript{2}\\
\small \textsuperscript{1}Shanghai American School, Shanghai, China 
\small \textsuperscript{2}Massachusetts General Hospital, Boston, MA}
\date{}

\begin{document}

\maketitle

\begin{abstract}
\textit{Objective:} To build a traceable multi-agent platform for breast-cancer liquid-biopsy research that supports biomarker discovery, evidence-grounded interpretation, controlled execution, and measured self-improvement of its own workflow.
\textit{Methods:} Biomarker-discovery agents coordinate modality-specific preprocessing, candidate screening, sparse modeling and evidence-based interpretation. A knowledge system combines a large-language-model wiki with keyword, vector and graph retrieval. A harness manages tools, execution state and provenance, and a regularized recursive self-improvement procedure revises the harness through bounded edits and measured retention. Patient-analysis agents link matched molecular and clinical records into research reports for physician review.
\textit{Results:} Three public cohorts comprising 821 samples or individual records were analyzed. Semantic-understanding accuracy of the knowledge system was 92.5\%, compared with approximately 66.5\% for a conventional knowledge base. In an expanded evaluation, self-improvement raised execution scores by approximately 23\%, 41\% and 47\% on development, held-out and out-of-distribution tasks. The patient-analysis agents produced 78 research reports from three matched-patient cohorts.
\textit{Conclusions:} The platform combines biomarker investigation, source-linked interpretation, traceable execution and measured workflow improvement, and establishes methodological feasibility. Clinical validity is not claimed.
\textit{Significance:} The work shows how agent workflows for biomedical discovery can be made inspectable and improvable without changing the analytical tools or the underlying model.
\end{abstract}


\section{Introduction}
Liquid biopsy makes tumor-associated molecular information accessible through blood, including cell-free DNA (cfDNA), circulating tumor cells (CTCs) and extracellular vesicles (EVs) \cite{keup2021,cristiano2019}. Nucleic acids, proteins and metabolites provide complementary material for breast-cancer biomarker research \cite{hart2017,gwark2021}. Identifying useful candidate markers requires more than generating measurements: the data must be processed appropriately, candidate signals assessed statistically, and their biological meaning examined against the available evidence. Assays differ in noise structure and quality requirements \cite{mestdagh2014,kirschner2011}, and sample heterogeneity and data partitioning affect feature selection and performance estimates \cite{desaire2022,tripod2024}.

We develop a platform that addresses three requirements of this process. First, a biomarker-discovery (BD) multi-agent framework coordinates preprocessing, candidate screening, panel construction and evaluation, and preserves the analytical records needed to assess each finding. Second, an enhanced medical knowledge system combines a large-language-model (LLM) wiki with keyword, vector and graph retrieval, so that candidate findings are linked to the passages that support their interpretation. Third, BioMed Harness manages skills, tool interfaces, execution state and analytical provenance, making the path from request to report inspectable.

The performance of an agent depends on both its model and the harness that governs how it uses information and tools. The harness determines which inputs are checked, how execution proceeds after a tool failure, how context is retained and whether findings reach a complete report. Recursive self-improvement (RSI) at this level uses feedback from current runs to propose and evaluate harness revisions while model weights stay fixed \cite{rrsi2026}. Repeated revision against a finite development set can, however, favor case-specific rules, track evaluation fluctuations, or add complexity without improving unseen tasks. We therefore adopt regularized RSI (RRSI), which constrains both candidate proposal and candidate selection \cite{rrsi2026}, and implement it for BioMed Harness with molecular tools and evaluation rules held fixed.

Finally, a patient-level analysis (PD) multi-agent framework links measurements from the same patient, generates modality-specific summaries, and examines agreement, conflict and supporting evidence in research reports for physician review.

The contributions are:
\begin{enumerate}
\item A BD multi-agent framework for liquid-biopsy preprocessing, candidate screening, panel construction and cohort-specific evaluation across three molecular data types.
\item A medical knowledge system combining an LLM wiki with keyword, vector and graph retrieval.
\item BioMed Harness, a platform for controlled and traceable agent operation, with skill and tool management, execution-state tracking, provenance and source-linked reporting.
\item An RRSI procedure that lets the platform improve its own execution strategy through bounded revisions and measured retention.
\item A PD framework that produced 78 structured research reports across three matched-patient cohorts.
\end{enumerate}

\section{Related Work}
\textit{Liquid-biopsy analysis.} Integrative analyses have combined blood analytes from the same patients \cite{keup2021}, and CTC transcriptome and paired tissue--serum miRNA studies extend this across molecular compartments \cite{ring2022,chan2013}. Assay-specific pipelines construct molecular features \cite{mirge2021,cfdnaanalyzer2026}, while statistical methods and workflow engines support downstream analysis and reproducible execution \cite{deseq2,snakemake}. Using several of these in one investigation still requires coordination of inputs, outputs and validation requirements.

\textit{Evidence-based interpretation.} Somatic-variant guidelines organize evidence levels \cite{amp2017}, and curated resources such as OncoKB and CIViC link alterations to interpretations \cite{oncokb2017,civic2017}. Retrieval-augmented generation connects LLM responses to external documents, and MIRAGE/MedRAG examines how corpus and retriever choices affect medical question answering \cite{medrag2024}. For molecular analysis, retrieved evidence must also be related to the measured analyte, disease setting and endpoint; the availability of a relevant publication does not establish the validity of a candidate biomarker.

\textit{Agent-based research systems.} Co-Scientist supports hypothesis generation \cite{coscientist2026}, and the harness literature examines how prompts, tools and control flow shape agent behavior \cite{rrsi2026}. This shifts attention from a single model response to a sequence of actions and its final deliverable. A biomedical application must consider both the methods that produce molecular results and the execution decisions that determine whether those results reach a report with their context intact.

\textit{Self-improvement.} Dream-RSI uses recorded discovery trees to evaluate alternative exploration policies through replay \cite{dreamrsi2026}. RRSI regularizes proposed harness edits and their retention to promote improvements that generalize beyond development tasks \cite{rrsi2026}. BioMed Harness provides a setting in which this question is examined for biomedical agents: can revisions to workflow coordination improve execution across assays and input conditions while preserving the analytical tools, evaluation rules and traceability of results?

\section{Platform Design}

\subsection{Overview and Scope Boundaries}
\begin{figure}[!t]
\centering
\includegraphics[width=0.65\columnwidth]{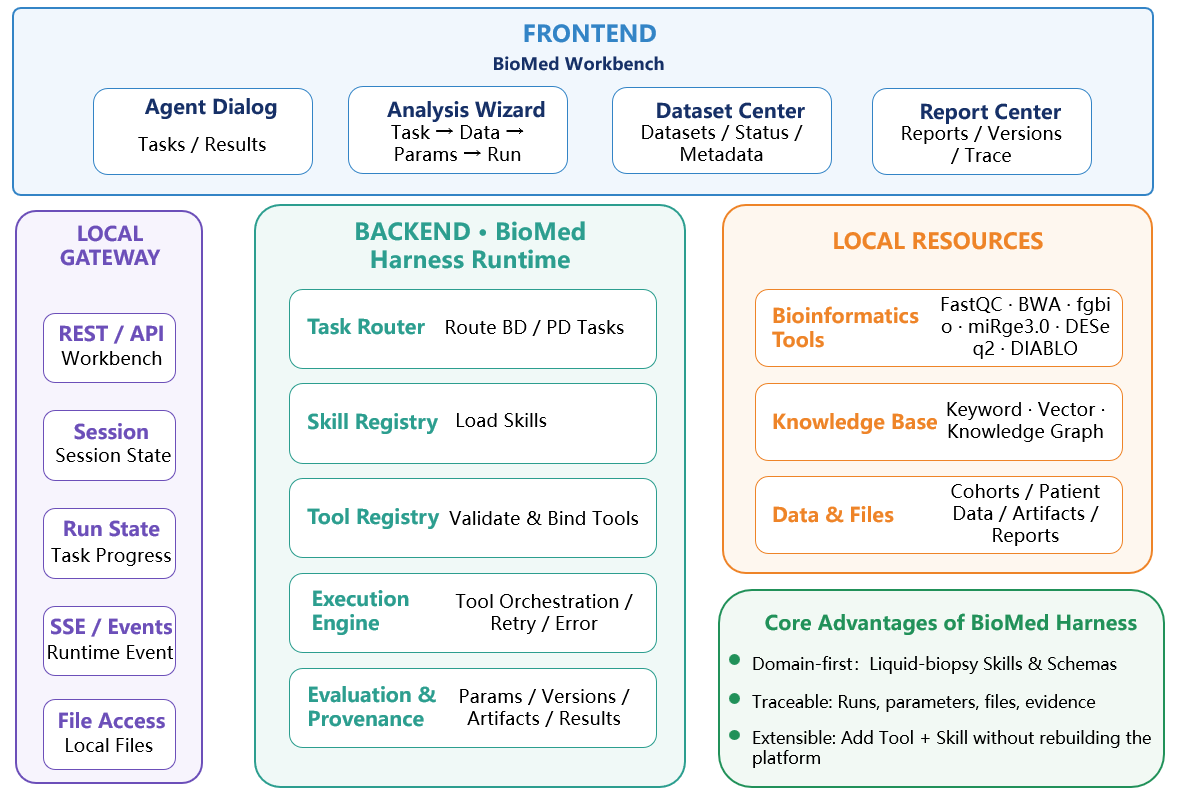}
\caption{Software architecture. The frontend connects through a local gateway to the BioMed Harness runtime, which coordinates analytical tools, evidence retrieval and data access for BD and PD workflows.}
\label{fig:arch}
\end{figure}

The platform has five connected functions (Fig.~\ref{fig:arch}). BD agents perform the scientific analysis of candidate liquid-biopsy biomarkers. The knowledge system supplies medical evidence for interpreting those candidates. BioMed Harness executes agent workflows through registered skills and tools, maintaining state, provenance and report delivery. The RRSI procedure uses execution feedback to revise and evaluate harness strategies. PD agents apply molecular analysis and knowledge-supported interpretation to measurements from the same patient. A browser-based workbench accepts BD or PD requests, supports dataset inspection and retrieves reports; a local gateway passes requests and execution updates to the harness, whose task router invokes the corresponding framework.

Each evaluation target is kept separate: BD reports candidate panels, cohort-specific discrimination and panel stability; the knowledge system is scored on semantic understanding; harness evaluation examines execution and provenance; PD reports patient-linked quality control (QC), measured findings, agreement or conflict, and evidence context. The platform distinguishes patient, sample and feature identifiers; BD uses sample-by-feature matrices with recorded cohort labels and split manifests, and PD links measurements by patient, analyte and sampling context.

All patient-level outputs are for research use only. The system may report measured findings, model scores and literature-supported associations. It does not treat candidate markers as diagnostic criteria, convert unvalidated model scores into absolute patient risk, or derive treatment decisions from knowledge-base retrieval.

\subsection{Biomarker-Discovery Agents}
\begin{figure}[!t]
\centering
\includegraphics[width=0.86\textwidth]{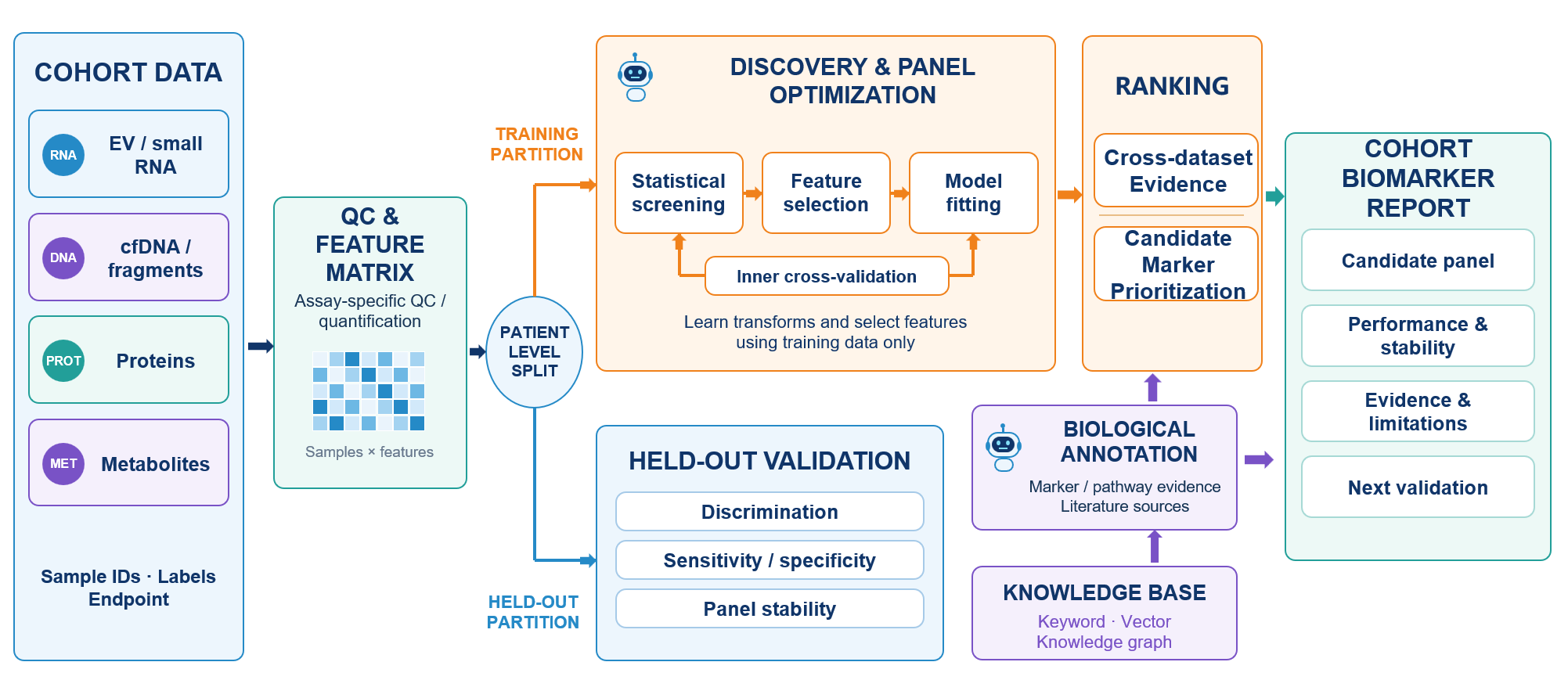}
\caption{Biomarker-discovery workflow. Solid arrows show the ordered analysis path from cohort inputs to the cohort report; dashed arrows indicate knowledge-base support for evidence ranking and source attribution. Feature selection and model tuning are fitted within training folds of the nested evaluation; predictions are evaluated on held-out sample partitions.}
\label{fig:bd}
\end{figure}

The BD pipeline accepts EV small-RNA, plasma-protein and serum-metabolite matrices (Fig.~\ref{fig:bd}). Each modality keeps its own preprocessing and QC contract before entering a common sample-by-feature modeling interface; the analysis then performs training-only feature selection and reports held-out performance and marker evidence.

\textit{Leakage control.} Feature filtering considers missingness, abundance, variance and pairwise correlation. Count-based screening uses DESeq2 with false-discovery-rate adjustment \cite{deseq2,bh1995}, and sparse panels are built with LASSO. In the strict nested evaluation of GSE270497, each inner training fold independently fits outcome-informed screening, scaling and model parameters; fitted transformations and selected features are applied to the corresponding validation samples without refitting, and split manifests verify disjoint training and validation sample identifiers. Archived single-split classifiers are evaluated separately.

\textit{Classification protocol.} For the cohort results in Section~\ref{sec:bd}, each repeated experiment used a stratified 80/20 train--test split. Up to 100 features were selected by absolute training-set $t$ statistics and evaluated with a HistGradientBoosting classifier of fixed configuration. Mean area under the receiver operating characteristic curve (AUC) and between-run standard deviation are reported, following TRIPOD+AI \cite{tripod2024}.

\begin{figure}[!t]
\centering
\includegraphics[width=0.92\textwidth]{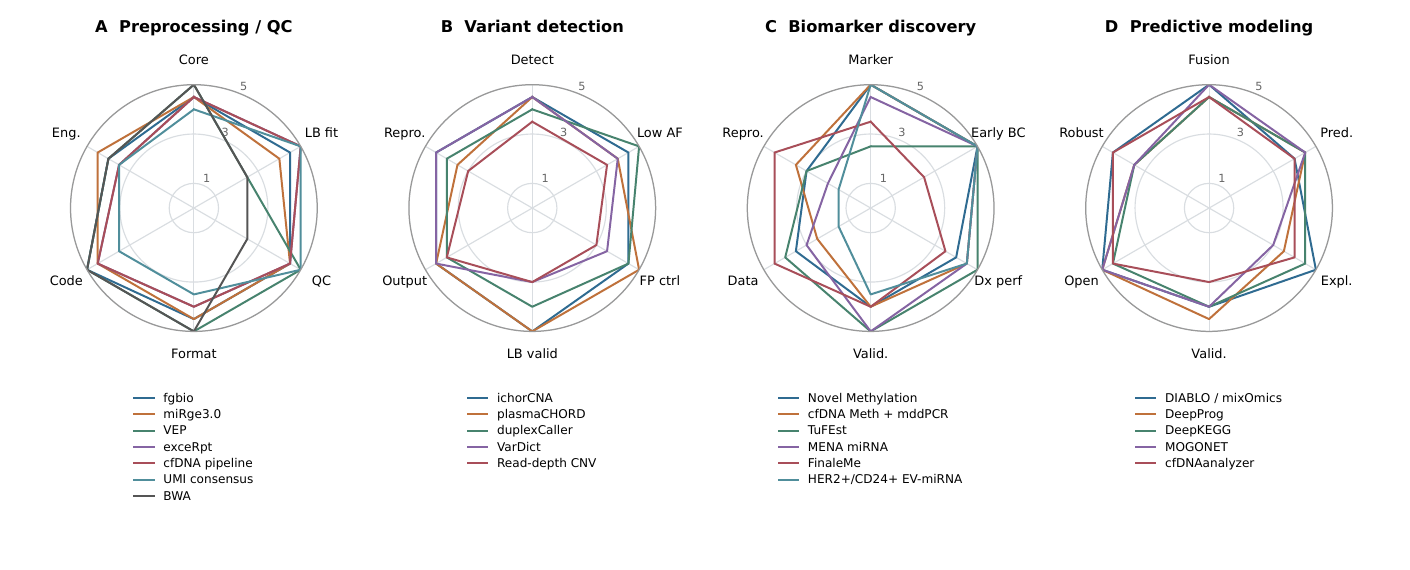}
\caption{Category-specific algorithm assessment. (A) Preprocessing, QC and annotation; (B) variant detection; (C) biomarker discovery; (D) predictive modeling. Profiles show six criteria on a 1--5 scale; criteria differ between categories and describe engineering suitability. LB: liquid biopsy; AF: allele frequency; FP: false positive.}
\label{fig:radar}
\end{figure}

\textit{Method selection and registration.} We assembled 23 candidate methods spanning preprocessing and annotation, variant detection, biomarker discovery and predictive modeling. Each was scored on six category-specific criteria (1--5), giving 138 recorded scores (Fig.~\ref{fig:radar}); scores guide engineering choices within a category and do not measure clinical performance. fgbio (27.0/30), miRge3.0 (26.5/30) and BWA (24.0/30) were adopted for unique-molecular-identifier (UMI) processing, small-RNA quantification and alignment, with the MENA workflow (24.0/30) and cfDNAanalyzer (24.5/30) supporting downstream marker analysis. DIABLO/mixOmics scored 27.5/30 but was not adopted, because the cohorts lacked the patient-matched multi-omics matrix it requires. Selected procedures are registered with explicit inputs, callable implementation, expected outputs and a validation task; the harness retains sample identifiers, parameters and output artifacts.

\subsection{Patient-Level Analysis Agents}
\begin{figure}[!t]
\centering
\includegraphics[width=0.86\textwidth]{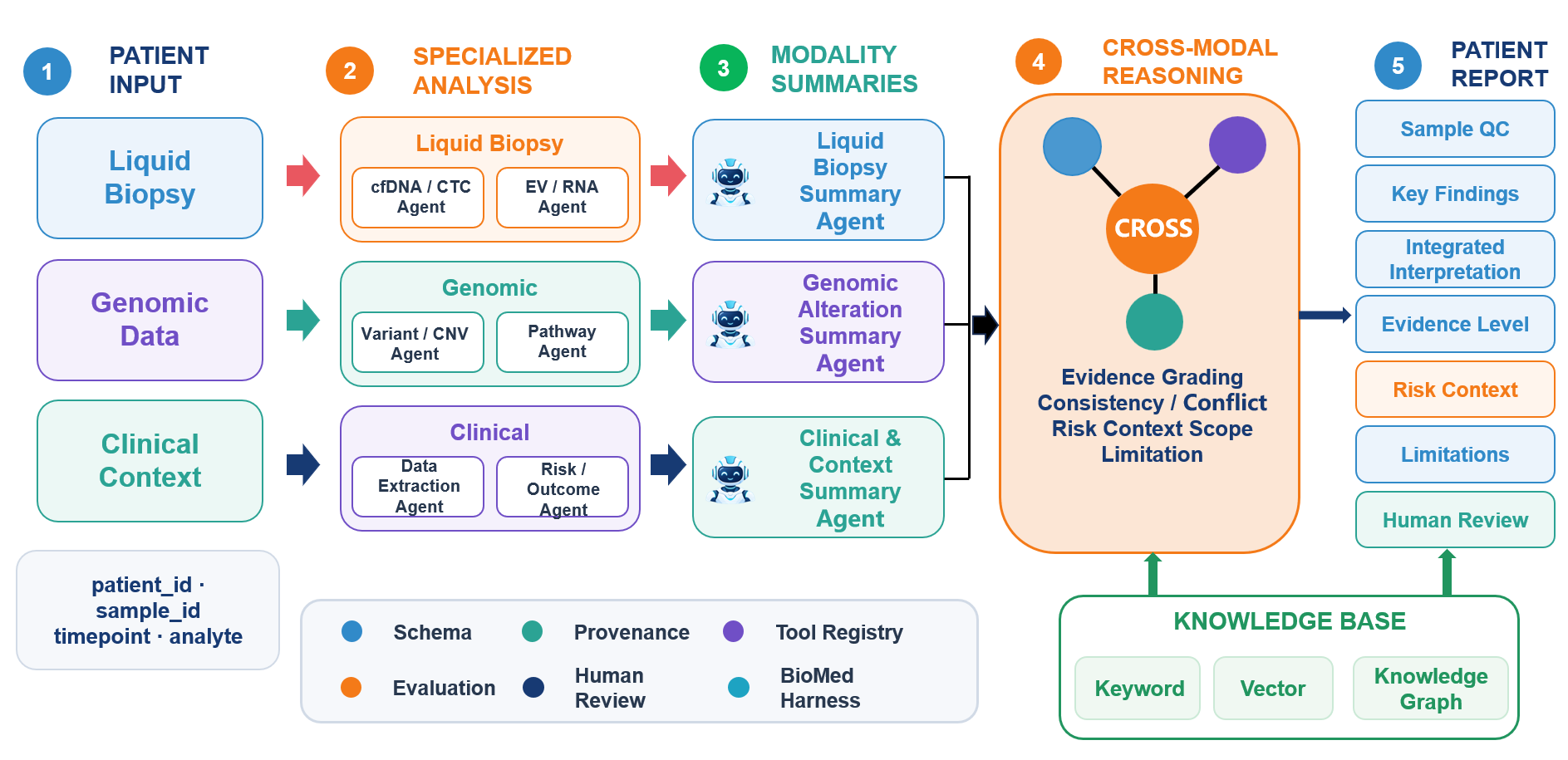}
\caption{Patient-level analysis workflow. Inputs remain in three independent branches through specialized analysis and modality summaries (steps 1--3) and converge at cross-modal reasoning (step 4). The knowledge base supplies supporting evidence; the research report retains measured findings, evidence, limitations and human review.}
\label{fig:pd}
\end{figure}

PD uses matched measurements from the same blood draw or paired tissue compartments. Liquid-biopsy, genomic and clinical inputs stay in independent branches through specialized analysis and modality summaries, and converge only at cross-modal reasoning (Fig.~\ref{fig:pd}). Patient, sample and collection-time identifiers are checked before findings are integrated.

\textit{Variant interpretation.} Detection quality and clinical significance are assessed separately. Interpretation uses AMP/ASCO/CAP tiers \cite{amp2017}, with variant-specific evidence checked in OncoKB and CIViC \cite{oncokb2017,civic2017}, cross-source reconciliation following curation and harmonization procedures \cite{danos2019,wagner2020}, oncogenicity classification standards \cite{horak2022} and ESCAT therapy-evidence categories \cite{escat2018}. Retrieved annotations keep their source and disease context. Model scores are reported with the evaluated cohort, endpoint and time horizon.

\textit{Evidence grading.} Detection reliability, biological or clinical relevance and actionability are recorded as separate attributes (Table~\ref{tab:evidence}). Concordant findings keep their measurement sources; discordant findings are reviewed against sampling time, tumor content, detection limits, tissue source, treatment history and knowledge-base version, and unresolved discrepancies stay in the report with possible explanations.

\begin{table}[!t]
\centering
\caption{Patient-Level Evidence Grading and Reporting Rules}
\label{tab:evidence}
\footnotesize
\setlength{\tabcolsep}{3pt}
\renewcommand{\arraystretch}{1.08}
\begin{tabular}{@{}>{\raggedright\arraybackslash}p{0.17\columnwidth}>{\raggedright\arraybackslash}p{0.42\columnwidth}>{\raggedright\arraybackslash}p{0.27\columnwidth}@{}}
\toprule
Level & Permitted statement & Human review \\
\midrule
E1 Detection fact & Detection results, values and assay status & Abnormal or borderline results \\
E2 Biological support & Association with a biological process or phenotype & When evidence conflicts \\
E3 Clinical association & Diagnostic or prognostic association in a defined population & Mandatory \\
E4 Therapy-related & Therapy-related evidence listed for professional evaluation & Mandatory sign-off \\
EU Uncertain & Uncertainty and information to supplement; not merged into a definitive conclusion & Mandatory \\
\bottomrule
\end{tabular}
\end{table}

\textit{Report generation and review.} Reports are rendered from structured JSON outputs and checked for numerical consistency, citation resolution and patient identity before human sign-off. Treatment-related, hereditary-risk and prognostic interpretations require professional review. Each report version stays associated with its input run, and reviewer edits are retained as annotations or structured differences.

\subsection{Knowledge System}
\begin{figure}[!t]
\centering
\includegraphics[width=0.92\textwidth]{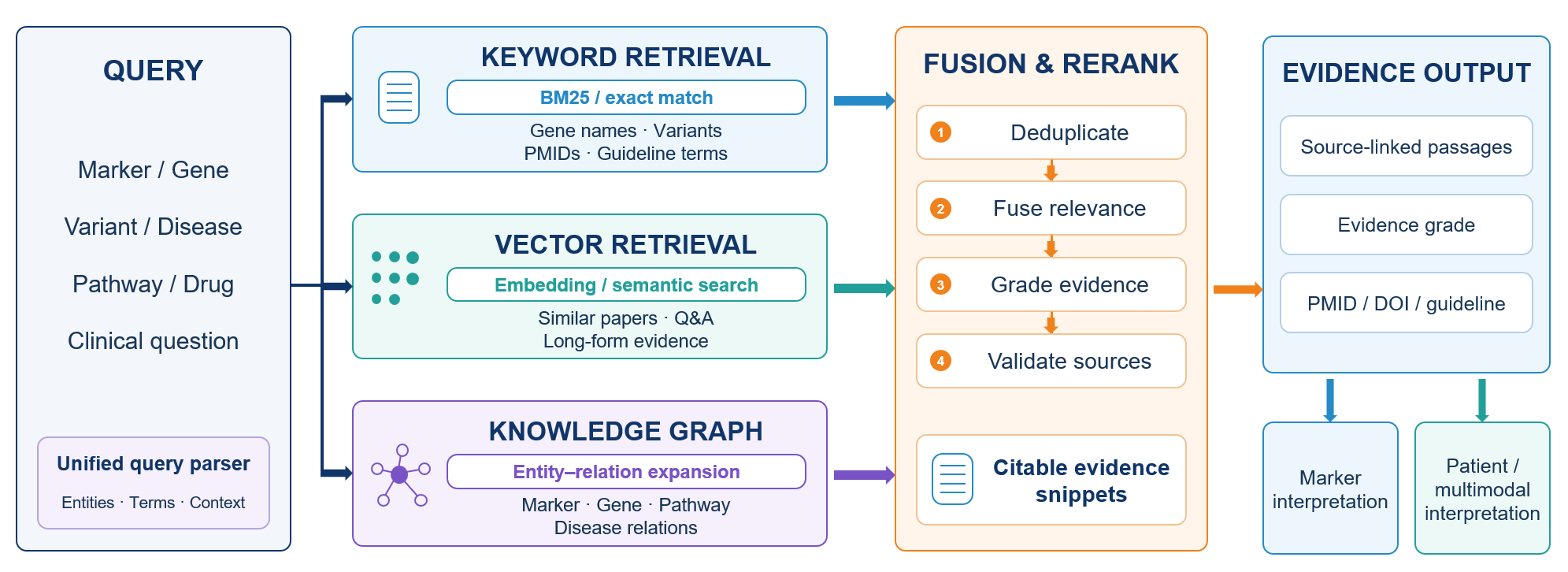}
\caption{Three-path retrieval and evidence fusion. Keyword, semantic and graph retrieval converge on source-linked evidence for marker and patient-level interpretation.}
\label{fig:retrieval}
\end{figure}

The knowledge base indexes documents as source passages, vector representations and entity relationships, with publication identifiers, versions and disease or specimen context attached to each record. A query parser identifies molecular entities, disease terms and the question asked, and the query then moves through three complementary routes (Fig.~\ref{fig:retrieval}) \cite{medrag2024,llmwiki}: keyword retrieval (BM25 or exact matching) for gene names, variants and guideline terms; vector retrieval for related passages whose wording differs from the query; and graph retrieval, which expands entity relationships from a marker to genes, pathways and diseases. Retrieved records enter evidence fusion and reranking, where duplicates are consolidated, relevance is assessed across routes, evidence is graded and source attribution is checked. The output keeps citable passages and identifiers so that a report statement can be checked against source text rather than a generated summary. Molecular values continue to come from analytical tool outputs; the knowledge base supplies attributed evidence for interpreting them.

\subsection{BioMed Harness}
BioMed Harness is the operational layer for the BD and PD frameworks. It manages registered skills and tools, checks input and output contracts, routes tasks and tracks execution state. Skills provide task-specific procedures, the tool registry binds callable implementations, and schemas define the fields passed between steps. Each tool call belongs to a run with a defined state: the harness records the inputs supplied, the tool invoked, its parameters and the output artifacts, and later steps consume these recorded outputs with the relevant sample and assay metadata. Report assembly keeps a measured value distinct from the evidence used to interpret it, so a report statement can be traced to both its measurement and its supporting source. The platform inventory contains 16 domain skills with defined tool contracts (Appendix~A). The execution records also supply the observations used by RRSI.

\subsection{Regularized Recursive Self-Improvement}
\begin{figure}[!t]
\centering
\includegraphics[width=0.86\textwidth]{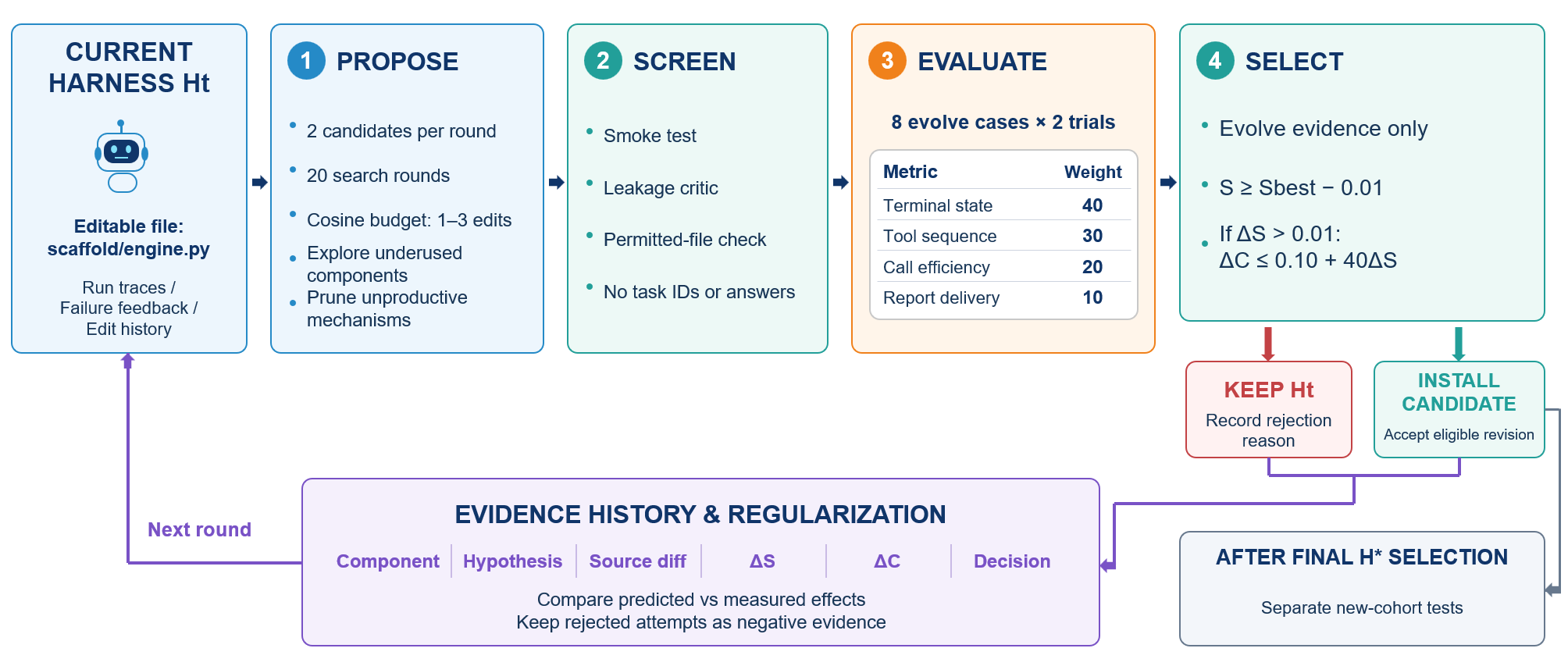}
\caption{Regularized recursive improvement of BioMed Harness. Candidate proposal, screening, evolve-set evaluation and constrained selection form the search loop. Rejection preserves the incumbent; acceptance installs an eligible improvement; both outcomes update the evidence history. Sealed held-out and new-cohort evaluations follow final-harness selection and do not feed back into search.}
\label{fig:rrsi}
\end{figure}

In RRSI, a run produces feedback, a proposer converts it into a candidate revision, and evaluation decides whether the revision becomes the harness used in the next round (Fig.~\ref{fig:rrsi}). The object of improvement is the executable policy around a frozen model: instructions, control flow, tool coordination and context management. Molecular-analysis tools, model weights, task assertions and evaluation rules remain fixed and are not modified to make a candidate appear more successful.

\textit{Proposal-side regularization.} An annealed edit budget limits how many mechanisms a candidate may change together. With $T=20$ rounds, $b_{\min}=1$ and $b_{\max}=3$,
\begin{equation}
b_t=\left\lceil b_{\min}+(b_{\max}-b_{\min})\frac{1+\cos(\pi t/T)}{2}\right\rceil .
\end{equation}
An evolution history records, for each evaluated edit, its component, hypothesis, source diff, score change, call-cost change and decision. Rejected mechanisms remain negative evidence, stalled search redirects effort toward underused components, and mechanisms without recent positive contribution become deletion targets.

\textit{Selection-side regularization.} A candidate must first pass smoke tests and critic review, which reject task-specific identifiers, embedded answers, benchmark-dependent branches and edits outside the permitted orchestration file. A frozen evaluator then scores complete runs; candidate code cannot assign its own score. Let $S_{\mathrm{best}}$ be the best observed evolve score, $\Delta S=S(H')-S(H_t)$ and $\Delta C=[C(H')-C(H_t)]/C(H_t)$. A candidate must satisfy
\begin{equation}
S(H')\ge S_{\mathrm{best}}-\delta,\qquad \Delta C\le 0.10+40\,\Delta S\ \text{ when }\Delta S>\delta,
\end{equation}
with $\delta=0.01$ retained as a fixed resolution floor because repeated unchanged-baseline trials had zero measured score variance. Analytical-tool calls serve as the resource proxy $C$ (the source RRSI study reports cost in policy tokens \cite{rrsi2026}). If no candidate qualifies, the current harness continues unchanged.

\textit{Execution score.} The evaluator assigns 40 points to final-state correctness, 30 to required tool order, 20 to call efficiency and 10 to report delivery; a hard assertion failure scores zero, and totals are normalized to $[0,1]$. The score measures workflow execution, not molecular classification performance.

\section{Experiments and Results}

\subsection{Datasets}
All datasets are public secondary-use data (Table~\ref{tab:data}). BD comprises 821 samples or records in total. PD uses three core matched-patient cohorts, plus one EV mRNA extension cohort that is not counted among the 78 core reports. Because the cohorts serve different validation purposes, their AUC values and the strength of their conclusions are not compared directly.

\begin{table}[!t]
\centering
\caption{Datasets Used for Biomarker Discovery (BD) and Patient-Level Analysis (PD)}
\label{tab:data}
\small
\renewcommand{\arraystretch}{1.08}
\begin{tabular}{@{}>{\raggedright\arraybackslash}p{0.22\columnwidth}>{\raggedright\arraybackslash}p{0.30\columnwidth}>{\raggedright\arraybackslash}p{0.38\columnwidth}@{}}
\toprule
Dataset & Material & Scale / role \\
\midrule
\multicolumn{3}{@{}l}{\textit{BD}} \\
GSE270497 \cite{geo270497,mo2024data} & Plasma EV small RNA & 180 samples (120 screening-positive, 60 screening-negative) \\
PXD028251 \cite{gwark2021} & Plasma proteome & 51 patients, 305 proteins (36 non-pCR, 15 pCR) \\
MTBLS424 \cite{hart2017} & Serum metabolome & 590 records; binary recurrence label \\
\multicolumn{3}{@{}l}{\textit{PD}} \\
ELIMA (PRJEB39331) \cite{keup2021} & cfDNA, CTC gDNA, CTC mRNA, EV mRNA, same draw & 26 HR+/HER2$-$ metastatic cases \\
GSE113890 \cite{ring2022} & CTC, lesion, blood transcriptomes & 21 stage IV cases + 5 healthy donors \\
GSE42128 \cite{chan2013} & Paired tissue and serum miRNA & 31 diagnostic-phase cases \\
GSE256523 & Serum EV mRNA (extension) & 30 samples; not in the 78 reports \\
\bottomrule
\end{tabular}
\end{table}

\subsection{Biomarker Discovery}\label{sec:bd}
\textit{GSE270497.} The median Pearson correlation between regenerated and author-provided count matrices was 0.998 across matched samples and features, indicating numerical preprocessing agreement. The repeated-split evaluation gave a mean AUC of 0.9264 (SD 0.0205). The discovery workflow progressively narrowed the feature set: 273 differentially abundant small RNAs were screened, LASSO shrank them to 25 candidate features, and a Top10 priority panel was formed from these candidates. Four of the Top10 candidates overlap the six qPCR-supported RNAs reported by the source study \cite{mo2024}. This priority panel is separate from the training-set feature ranking used in the classifier.

\begin{figure}[!t]
\centering
\includegraphics[width=0.65\columnwidth]{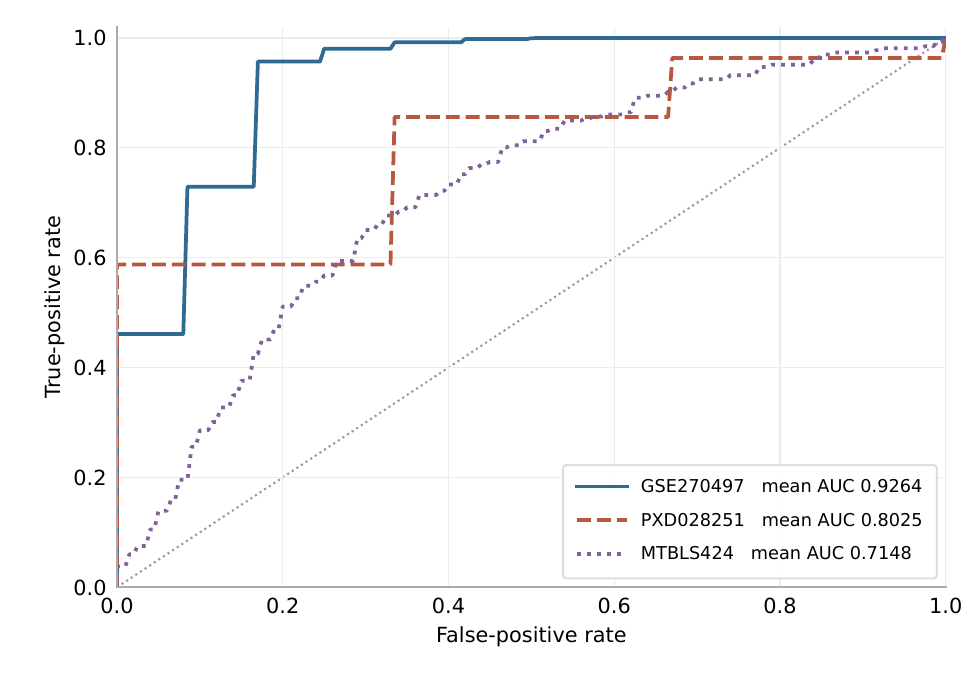}
\caption{Mean ROC profiles from the repeated-split BD evaluation. The legend gives the mean run-wise AUC per cohort. Standard deviations are reported in the text and are not confidence intervals. The dotted diagonal indicates chance.}
\label{fig:roc}
\end{figure}

\textit{Proteomic and metabolomic cohorts.} With non-pCR as the positive label, PXD028251 gave a mean AUC of 0.8025 (SD 0.0695). MTBLS424 gave 0.7148 (SD 0.0092) for binary recurrence (Fig.~\ref{fig:roc}).

\textit{Comparison with published studies.} We compared BD results with one published study per data type (Table~\ref{tab:ref}): an exosomal small-RNA model with six RNAs \cite{mo2024}, a plasma-protein reference for neoadjuvant response \cite{gwark2021}, and a serum-metabolomics recurrence model \cite{hart2017}. The published and BD figures come from different protocols, partitions and classifiers; the table does not establish superiority.

\begin{table}[!t]
\centering
\caption{Cohort-Specific AUC of BD and Published Reference Methods. Mo et al.: test-set AUC. Gwark et al.: median AUC of the model built on the three selected proteins (ENG, MBL2, P4HB) over repeated splits. Hart et al.: validation AUC of the initial random-forest model, and test AUC of the optimized model on a cohort filtered to at least 6 years of follow-up, with lactate signal removed.}
\label{tab:ref}
\footnotesize
\begin{tabular}{@{}l@{\hspace{4pt}}l@{\hspace{4pt}}c@{\hspace{4pt}}c@{}}
\toprule
Dataset & Reference & Ref.\ AUC & BD AUC \\
\midrule
GSE270497 & Mo et al.\ \cite{mo2024} & 0.97 & 0.93 \\
\shortstack[l]{PXD028251\\(selected features)} & Gwark et al.\ \cite{gwark2021} & 0.86 & 0.80 \\
MTBLS424 & Hart et al.\ \cite{hart2017} & 0.63 & 0.71 \\
\shortstack[l]{MTBLS424\\(filtered cohort)} & Hart et al.\ \cite{hart2017} & 0.75 & 0.71 \\
\bottomrule
\end{tabular}
\end{table}

\textit{Strict nested evaluation of GSE270497.} The reselected workflow was evaluated on the 180 samples and 1,339 input features with 100 repetitions of five-fold cross-validation, using five inner folds within each outer training set. Each inner training fold independently ran DESeq2 screening, scaling and L1-regularized logistic regression; the penalty was chosen by inner validation log loss and the decision threshold by inner out-of-fold Youden's $J$, and all 3,000 DESeq2 tasks completed. Across the 100 repetitions the pooled out-of-fold AUC had a mean of 0.9745 (SD 0.0068) and ranged from 0.9501 to 0.9861, with an empirical 2.5th--97.5th repeated-split percentile range of 0.9594--0.9844. The archived single-split AUC of 0.9856 lies at the upper end of this distribution, above the 97.5th percentile. These percentiles describe partition variability and are not an independent-population confidence interval. Each model retained 17.0 nonzero features on average (SD 2.0); hsa-miR-484 was selected in 500 of 500 outer models and hsa-miR-9-5p in 484 of 500. The earlier single-split classifier, recalculated from archived prediction files, used 17 features on 60 positive and 30 negative held-out samples, and therefore differs from both the 25-candidate discovery illustration and the nested workflow.

\textit{Measurement batch in PXD028251.} Acquisition dates separated 35 samples (25 non-pCR, 10 pCR) from 16 (11 non-pCR, 5 pCR); outcome composition did not differ (Fisher exact $p=1.000$). Measurement distributions nevertheless differed by acquisition date: PC1 and PC2 each had Mann--Whitney $p<0.0001$, and 126 of 305 proteins had Benjamini--Hochberg (BH) adjusted $q<0.05$. This context applies when interpreting the cohort-level AUC.

\subsection{Patient-Level Analysis}
\textit{Report production.} The three core cohorts produced 78 patient-level research reports: 26 (ELIMA), 31 (GSE42128) and 21 (GSE113890).

\textit{ELIMA cross-analyte agreement.} The paper-matched variant layer comprised 51 patient--analyte records (28 cfDNA, 23 CTC gDNA). Of 26 patients, ten were positive in both analytes, six in cfDNA only, four in CTC gDNA only and six in neither; 14 patient--variant records were detected in both analytes across the ten double-positive patients. Patient-level agreement was $\kappa=0.217$ (patient-bootstrap 95\% CI $-0.168$ to $0.601$), and the exact McNemar test of marginal positivity gave $p=0.7539$. Excluding germline-positive annotations gave $\kappa=0.012$. These findings describe report-level concordance and are interpreted descriptively.

\textit{UMI family-size sensitivity.} Original paired-end FASTQ files for 52 samples from 26 patients were reprocessed with fgbio, BWA, SAMtools and LoFreq at minimum family sizes of one, two and three reads. All 156 sample--setting runs completed. The median singleton-family proportion was 0.364. Median consensus molecules fell from 256,654 to 38,071 and 27,205, PASS candidate records from 29,738 to 14,579 and 11,386, and recovery of the 51 archived patient--analyte records from 51/51 to 30/51 and 23/51. Archived-record recovery is not an estimate of true variant-detection accuracy. At the one-read setting, 29,630 independent candidates passed the predefined criteria; all 26 patients were positive in both analytes, leaving no variation for an informative agreement estimate, and the analytes overlapped at 5,595 patient--variant records.

\textit{Tissue--serum direction analysis (GSE42128).} Using a documented within-compartment reference and $|\Delta|\ge 0.5$, the reanalysis found 21--62 concordant miRNAs per patient (median 39; total 1,186). Cross-patient pairing and within-patient direction-permutation null means were approximately 1,171 and 1,175, and no patient reached BH-adjusted $q<0.05$ (0/31 in each analysis). Concordant directions are therefore reported as descriptive findings.

\begin{figure}[!t]
\centering
\includegraphics[width=0.65\columnwidth]{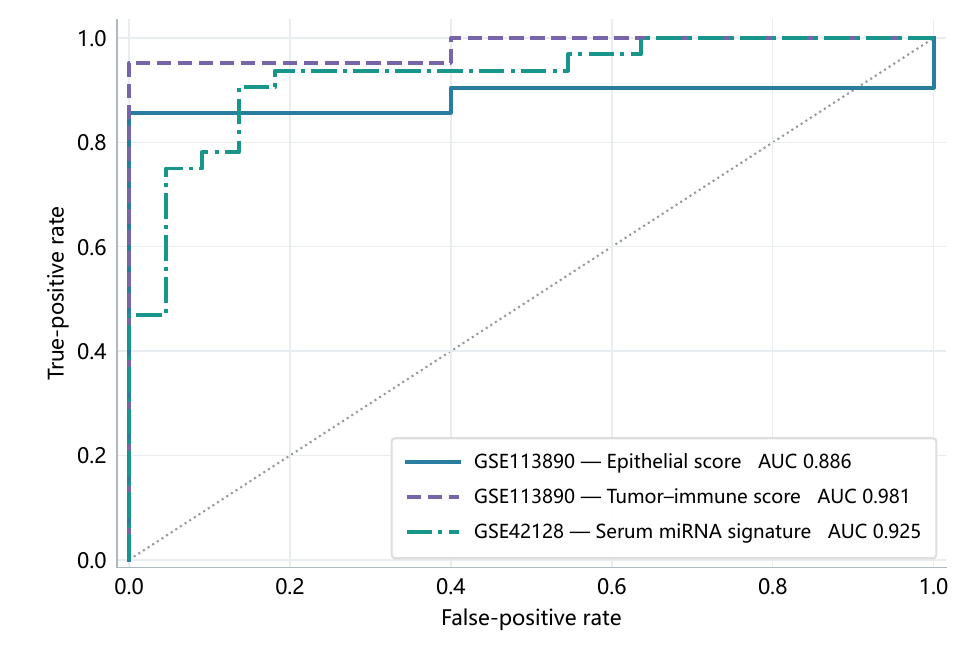}
\caption{ROC curves for report-derived scores against dataset case/control labels (case = 1). GSE113890: epithelial and tumor--immune scores (21 cases, five healthy donors). GSE42128: serum miRNA signature (32 cases, 22 controls). The dotted diagonal indicates chance.}
\label{fig:pdroc}
\end{figure}

\textit{Report-derived molecular scores.} We evaluated whether molecular measurements retained in the reports preserve case/control information (Table~\ref{tab:pd}, Fig.~\ref{fig:pdroc}). In GSE113890, an epithelial score (A1) averaged the peripheral-blood $\log_2$(CPM+1) values of KRT19, KRT8, KRT18, EPCAM and CDH1, and a tumor--immune score (A2) subtracted the mean of PTPRC, CD19, CD3E and B2M from the mean of 19 tumor/luminal genes; both used 21 cases and five healthy donors. In GSE42128, a serum signature was built from the 31 reports: a miRNA entered the upregulated (downregulated) set if it appeared at least five times in that direction and more often than in the opposite direction, giving 30 upregulated and 38 downregulated miRNAs. The signature score (A3) was the upregulated-set mean minus the downregulated-set mean, evaluated on 32 breast-cancer cases and 22 healthy controls; a leave-one-report analysis (A4) rebuilt the signature from the remaining 30 reports for each of the 31 reports. These summaries use no fitted classifier coefficients. AUC uncertainty for A1--A3 used 2,000 subject-level bootstrap resamples.

A1 gave AUC 0.8857 (95\% CI 0.733--1.000; mean score 1.8711 in cases and 0.3536 in controls; 18 of 21 cases exceeded the highest control). A2 gave 0.9810 (0.913--1.000), indicating stronger separation after immune-background subtraction, with a small control group limiting precision. A3 gave 0.9247 (0.844--0.984); A4 gave a mean AUC of 0.9260 (2.5th--97.5th percentile 0.917--0.937; range 0.916--0.939), which tests dependence on individual reports rather than providing held-out predictions. As a supervised reference, XGBoost was trained on 200 repeated stratified 70/30 splits using 26$\times$5, 26$\times$23 and 54$\times$68 (samples$\times$features) matrices. Report-derived scores were numerically close to these references.

\begin{table}[!t]
\centering
\caption{AUC of Report-Derived Scores and XGBoost References}
\label{tab:pd}
\small
\begin{tabular}{@{}lcc@{}}
\toprule
Dataset / score & XGBoost & Report-derived \\
\midrule
GSE113890, epithelial (A1) & 0.89 & 0.89 \\
GSE113890, tumor--immune (A2) & 1.00 & 0.98 \\
GSE42128, serum miRNA (A3) & 0.97 & 0.92 \\
\bottomrule
\end{tabular}
\end{table}

\subsection{Knowledge System}
\begin{figure}[!t]
\centering
\includegraphics[width=0.85\textwidth]{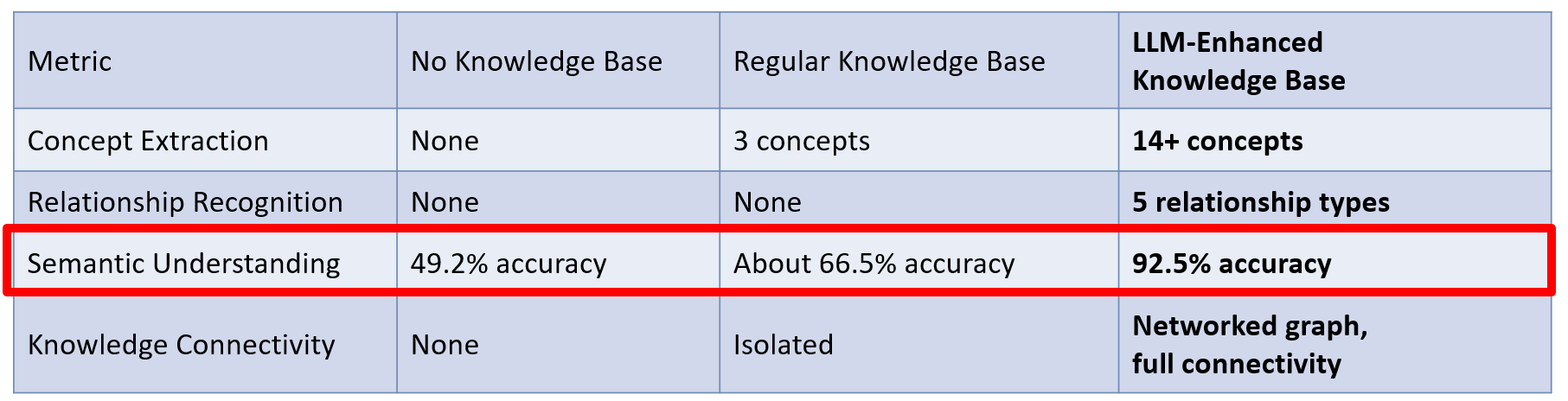}
\caption{Knowledge-base comparison: the LLM wiki with three-path retrieval versus a conventional knowledge base and analysis without a knowledge base, on concept extraction, relationship recognition and semantic understanding.}
\label{fig:kb}
\end{figure}

We evaluated the LLM wiki with three-path retrieval using MedQA questions and the Textbooks corpus. Table~\ref{tab:medqa} reproduces the six published MedQA-US results for the Textbooks corpus from MIRAGE/MedRAG \cite{medrag2024}; these question-answering accuracies and our semantic-understanding scores are separate measures. The enhanced system identified more than fourteen concepts and five relationship types, versus three concepts and no recognized relationships for the conventional knowledge base (Fig.~\ref{fig:kb}). Semantic-understanding accuracy reached 92.5\%, compared with approximately 66.5\% for the conventional knowledge base and 49.2\% without a knowledge base, gains of approximately 26.0 and 43.3 percentage points.

\begin{table}[!t]
\centering
\caption{Published MedQA-US Accuracy of GPT-3.5 with MedRAG and the Textbooks Corpus \cite{medrag2024}}
\label{tab:medqa}
\footnotesize
\setlength{\tabcolsep}{2pt}
\begin{tabular}{@{}lcccccc@{}}
\toprule
Retriever & BM25 & Contriever & SPECTER & MedCPT & RRF-2 & RRF-4 \\
\midrule
Accuracy (\%) & 66.54 & 67.16 & 67.40 & 66.22 & 65.91 & 66.06 \\
\bottomrule
\end{tabular}
\end{table}

\subsection{Harness Self-Improvement}
\textit{Setup.} The original search used 20 rounds with two independent candidate proposals per round. Analyst, proposer, critic and digester roles used openai/gpt-oss-120b through a local OpenAI-compatible endpoint. Eight evolve tasks (E01--E08) used public omics matrices from GSE270497, GSE42128, PXD028251, MTBLS424, CPTAC-BRCA and DELFI cfDNA under clean, missing-label, ID-mismatch, paired-block, transient-failure, persistent-failure, missing-block and unsupported-mode conditions. Four sealed held-out tasks (H01--H04) used GSE256523 and GSE113890, and six new-cohort tasks used GSE111842; held-out and new-cohort tasks were evaluated only after selection (Appendix~B).

\textit{Retained updates.} The baseline harness scored 0.7438 on the eight development tasks (Table~\ref{tab:rrsi}). A temporary preprocessing error could stop a recoverable run, and reports from incomplete inputs could receive an unqualified status. The first retained update allowed one preprocessing retry; the second connected failure reporting and input disclosure to the final status; the third restricted retry to transient errors, removing an ineffective call after persistent failures. The final harness satisfied all eight development tasks. Mean tool use rose from 3.0 to 3.5 calls per task, and conditional retry lowered calls from 3.625 (unconditional retry) to 3.500.

\begin{table}[!t]
\centering
\caption{Cumulative Execution Improvements in the 20-Round Search. Rounds 1--3 label successive retained-update stages, not proposal-round indices; each stage includes earlier updates. Scores are means over eight development tasks.}
\label{tab:rrsi}
\small
\renewcommand{\arraystretch}{1.08}
\begin{tabular}{@{}>{\raggedright\arraybackslash}p{0.13\columnwidth}>{\raggedright\arraybackslash}p{0.55\columnwidth}r@{}}
\toprule
Stage & Retained change & Evolve score \\
\midrule
Baseline & Original execution policy & 0.74 \\
Round 1 & Allow one preprocessing retry & 0.83 \\
Round 2 & Deliver qualified reports; disclose available inputs & 0.99 \\
Round 3 & Restrict retry to transient errors & 1.00 \\
\bottomrule
\end{tabular}
\end{table}

\textit{Held-out and new-cohort evaluation.} On four sealed tasks, the mean execution score rose from 0.8000 to 1.0000 without more tool use (4.0 calls per task); both incomplete-input reports disclosed the missing information and received a qualified status. On six GSE111842 tasks (complete input, two missing-compartment inputs, persistent failure, transient failure and identifier mismatch), the baseline mean was 0.6583 and the selected harness satisfied all six. These results concern report handling under specified input conditions, not molecular performance.

\textit{Ablation.} Table~\ref{tab:abl} separates the contributions. Retry alone raised the development score but left the held-out score unchanged; reporting and disclosure alone corrected incomplete-input reports but did not fully recover the transient-failure task; their combination covered both, and conditional retry then removed the ineffective repeat. Raising the call budget changed neither score nor observed calls, whereas lowering it interrupted the five-call recovery path.

\begin{table}[!t]
\centering
\caption{Post-Run Ablation of Recovery and Reporting Behaviors. Dev.: eight development tasks; Held-out: four sealed tasks; New cohort: six GSE111842 tasks; Calls: mean analytical-tool calls per development task. Rows are post-run policy comparisons.}
\label{tab:abl}
\small
\setlength{\tabcolsep}{3pt}
\begin{tabular}{@{}>{\raggedright\arraybackslash}p{0.40\columnwidth}cccc@{}}
\toprule
Policy & Dev. & Held-out & New cohort & Calls \\
\midrule
Baseline & 0.74 & 0.80 & 0.66 & 3.00 \\
Retry only & 0.83 & 0.80 & 0.77 & 3.50 \\
Reporting only (no retry) & 0.94 & 1.00 & 0.92 & 3.25 \\
Reporting + unconditional retry & 0.99 & 1.00 & 0.98 & 3.63 \\
\textbf{Selected (transient-only retry)} & \textbf{1.00} & \textbf{1.00} & \textbf{1.00} & \textbf{3.50} \\
\bottomrule
\end{tabular}
\end{table}

\begin{figure}[!t]
\centering
\includegraphics[width=0.86\textwidth]{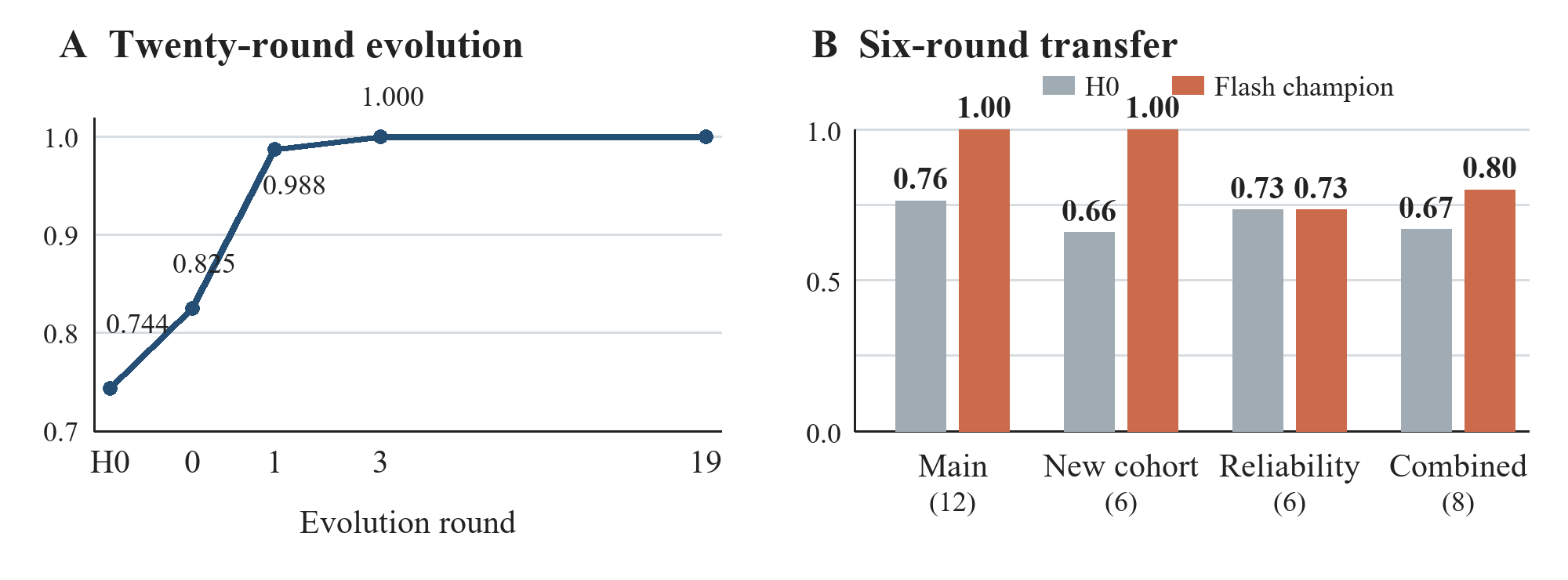}
\caption{Execution gains during search and subsequent evaluation. (A) Mean evolve execution score during the original 20-round search. (B) Baseline and evolved Flash execution scores on the original 12 tasks and the new-cohort (N), result-reliability (R) and combined-challenge (Q) families; three independent Flash seeds produced identical family means.}
\label{fig:rrsires}
\end{figure}

\textit{Independent searches and transfer.} Nine independent six-round searches also retained changes to retry, failure reporting and input disclosure: three in one update, six with a later transient-only refinement. All nine final versions satisfied the eight development tasks and averaged 3.5 calls per task. In an independent Flash evaluation of the original 12 tasks, final-state correctness accounted for the largest component gain (20.0 to 40.0 of 40; tool order 27.9 to 30.0 of 30; call efficiency 20.0 of 20 unchanged; report delivery 8.3 to 10.0 of 10). Scores improved on the original tasks, the new cohort and a combined-challenge family, but a result-reliability family (weak discrimination or unstable marker panels) stayed at 0.73 before and after (Fig.~\ref{fig:rrsires}); the retained changes therefore improved recovery and report handling without improving those reliability decisions.

\textit{Expanded evaluation with Qwen.} An expanded benchmark used ten LB-Evolve tasks for generating and selecting revisions, eight ID-held-out tasks and seven out-of-distribution GSE111842 tasks. Two independent six-round Qwen searches retained the retry, failure-reporting and input-disclosure changes in the first round. Relative execution-score gains were 23\%, 41\% and 47\% on development, held-out and out-of-distribution tasks (Table~\ref{tab:qwen}); each gain is relative to its own family baseline, and the families are not pooled with the original 8/4-task suite.

\begin{table}[!t]
\centering
\caption{Relative Execution-Score Gains in the Expanded Qwen Evaluation}
\label{tab:qwen}
\small
\begin{tabular}{@{}lccr@{}}
\toprule
Task family & Tasks & Role & Gain \\
\midrule
LB-Evolve & 10 & Selection and retention & 23\% \\
ID-held-out & 8 & Assessment after selection & 41\% \\
GSE111842 & 7 & Out-of-distribution & 47\% \\
\bottomrule
\end{tabular}
\end{table}

\section{Discussion}
\textit{Principal findings.} The platform carries liquid-biopsy analyses from raw inputs to source-linked reports under explicit execution control, and its harness can improve its own execution policy under fixed tools and evaluation rules. The retained RRSI changes concern when to retry, when to stop and how to deliver a qualified report; they improved execution on held-out and new-cohort tasks, while a result-reliability family did not improve. The knowledge system raised semantic-understanding accuracy over a conventional knowledge base, and the PD agents produced 78 reports whose molecular summaries retained case/control information within the evaluated cohorts. The PD molecular comparisons also showed modest agreement: ELIMA $\kappa=0.217$, and no GSE42128 patient with BH-adjusted $q<0.05$. The reports preserve such results as observed.

\textit{Limitations.} The evidence supports methodological feasibility, not clinical validity. (i) The datasets are public, small and cohort-specific, with no external validation; PXD028251 has 51 patients and a measurable acquisition-date batch effect, and the MTBLS424 AUC is 0.7148. (ii) Comparisons with published studies use different protocols and partitions. (iii) No comparison against a conventional non-agent pipeline, a single-LLM baseline or standard retrieval is reported, and the knowledge-system score is a semantic-understanding measure defined for this study rather than a standard benchmark accuracy. (iv) The serum signature is derived from the 31 reports, the healthy-reference samples used during report generation are included in its evaluation, and the control group in GSE113890 has five donors. (v) RRSI is evaluated on engineered execution tasks over real data matrices, scores reach 1.0 on the original suites, and the ablation isolates recovery and reporting behaviors rather than each regularization component. (vi) No independent expert audit of the 78 reports is reported. (vii) The execution score does not measure molecular accuracy.

\textit{Future work.} Planned work includes independent-cohort validation of marker panels with locked preprocessing and feature selection, longitudinal multi-analyte patient assessment with expert review, evaluation of retrieval relevance and evidence-grade agreement on marker and patient-specific questions, and RRSI across additional assay interfaces under matched budgets.

\section{Conclusion}
BioMed Harness combines biomarker discovery, source-linked medical evidence, traceable execution and regularized self-improvement for breast-cancer liquid-biopsy research. Across three BD cohorts, three matched-patient cohorts and several execution task families, the platform completed analyses, produced 78 research reports, and improved its own recovery, disclosure and reporting behavior through bounded revisions with measured retention. These findings support harness-level self-improvement as a way to strengthen agent execution through transferable control rules, within the limits stated above.

\appendix

\section{Tool and Skill Inventory}
The platform inventory contains 16 domain skills; the core modules, main inputs, deterministic computation and outputs are listed in Table~\ref{tab:tools}. A callable preprocessing or analysis module is distinguished from an inferred clinical capability, which still requires validation in the target setting.

\begin{table}[!t]
\centering
\caption{Core Modules, Inputs, Deterministic Computation and Outputs}
\label{tab:tools}
\footnotesize
\renewcommand{\arraystretch}{1.05}
\begin{tabular}{@{}>{\raggedright\arraybackslash}p{0.21\textwidth}>{\raggedright\arraybackslash}p{0.19\textwidth}>{\raggedright\arraybackslash}p{0.28\textwidth}>{\raggedright\arraybackslash}p{0.24\textwidth}@{}}
\toprule
Module & Main input & Deterministic computation & Main output \\
\midrule
liquid-biopsy-preprocessing-agent & File manifest, sample metadata & Data type identification, integrity check, routing & Data type, missing fields, downstream workflow \\
liquid-biopsy-small-rna-preprocess & Small-RNA FASTQ & FastQC, Cutadapt \cite{cutadapt2011}, miRge3.0 \cite{mirge2021} & QC, miRNA counts/TPM \\
exrna-preprocess / exSEEK & exRNA FASTQ or count matrix & Multi-biotype quantification, normalization, feature selection & exRNA expression matrix, candidate features \\
liquid-biopsy-dna-preprocess & cfDNA/CTC DNA paired-end FASTQ & FastQC, BWA \cite{bwa2009}, SAMtools \cite{samtools2009} & Sorted, indexed BAM; QC metrics \\
liquid-biopsy-umi-cfdna-preprocess & UMI paired-end FASTQ & UMI extraction, grouping, consensus or deduplication & Consensus-ready BAM, UMI QC \\
cfdna-analyzer & cfDNA BAM & CNA, fragment length, coverage and end features \cite{cfdnaanalyzer2026} & Sample$\times$feature matrix \\
variant analysis & BAM, reference genome & Low-frequency calling (LoFreq \cite{lofreq2012}), annotation, filtering & VCF, annotation table, candidate variants \\
deseq2-biomarker-agent & Count matrix, group information & Negative binomial model, multiple-testing correction & Differential feature table \\
liquid-biopsy-marker-discovery-agent & Training/external matrix & LASSO panel construction, held-out evaluation & Candidate panel, coefficients, AUC \\
marker-ranking-report & Candidate markers, knowledge-base evidence & Score normalization, evidence retrieval, template rendering & Top-$N$ list, evidence table, report \\
\bottomrule
\end{tabular}
\end{table}

\section{RRSI Task and Search Details}
Eight evolve tasks: E01 clean EV small-RNA classification (GSE270497); E02--E04 on GSE42128 (missing labels, mismatched sample IDs, paired tissue and serum blocks); E05 transient and E06 persistent preprocessing failure (PXD028251, MTBLS424); E07 one of four omics blocks omitted (CPTAC-BRCA); E08 unsupported cfDNA mode (DELFI). Task-specific terminal states, required calls, prohibited calls and report modes were fixed before evolution. Sealed tasks: H01--H02 on GSE256523 (complete; missing labels), H03--H04 on GSE113890 (complete; one compartment missing); their files and traces were hidden from all roles during the run. Family N used GSE111842 CTC, tumor and peripheral-blood matrices (six tasks); two further post-run families tested result reliability (Family R, six cases) and combined reliability and workflow difficulty (Family Q, eight cases), requiring a limited conclusion when cross-fold panel Jaccard was below 0.30 or cross-validated AUC below 0.65. The failure scenarios are controlled perturbations over real expression matrices. Eleven deterministic harness variants (H0, three cumulative edit subsets, the retained harness, two leave-out variants and four hand-written comparison policies) were scored by the same evaluator on the evolve and sealed cases. The prespecified post-run PolicyScore was $0.4\,S_{\mathrm{evolve}}+0.6\,S_{\mathrm{held\text{-}out}}$ and was not a selection objective. Two deterministic trials were run per original case.

\bibliographystyle{unsrtnat}
\bibliography{refs}

@article{keup2021, author={C. Keup and V. Suryaprakash and S. Hauch and others}, title={Integrative statistical analyses of multiple liquid biopsy analytes in metastatic breast cancer}, journal={Genome Med.}, volume={13}, pages={85}, year={2021}}

@article{cristiano2019, author={S. Cristiano and A. Leal and J. Phallen and others}, title={Genome-wide cell-free {DNA} fragmentation in patients with cancer}, journal={Nature}, volume={570}, number={7761}, pages={385--389}, year={2019}}

@article{hart2017, author={C. D. Hart and A. Vignoli and L. Tenori and others}, title={Serum metabolomic profiles identify {ER}-positive early breast cancer patients at increased risk of disease recurrence in a multicenter population}, journal={Clin. Cancer Res.}, volume={23}, number={6}, pages={1422--1431}, year={2017}}

@article{gwark2021, author={S. Gwark and H. S. Ahn and J. Yeom and others}, title={Plasma proteome signature to predict the outcome of breast cancer patients receiving neoadjuvant chemotherapy}, journal={Cancers}, volume={13}, number={24}, pages={6267}, year={2021}}

@article{mestdagh2014, author={P. Mestdagh and N. Hartmann and L. Baeriswyl and others}, title={Evaluation of quantitative {miRNA} expression platforms in the {microRNA} quality control ({miRQC}) study}, journal={Nat. Methods}, volume={11}, number={8}, pages={809--815}, year={2014}}

@article{kirschner2011, author={M. B. Kirschner and S. C. Kao and J. J. Edelman and N. J. Armstrong and M. P. Vallely and N. van Zandwijk and G. Reid}, title={Haemolysis during sample preparation alters {microRNA} content of plasma}, journal={PLoS ONE}, volume={6}, number={9}, pages={e24145}, year={2011}}

@article{desaire2022, author={H. Desaire}, title={How ({N}ot) to generate a highly predictive biomarker panel using machine learning}, journal={J. Proteome Res.}, volume={21}, number={9}, pages={2071--2074}, year={2022}}

@article{tripod2024, author={G. S. Collins and K. G. M. Moons and P. Dhiman and others}, title={{TRIPOD+AI} statement: Updated guidance for reporting clinical prediction models that use regression or machine learning methods}, journal={BMJ}, volume={385}, pages={e078378}, year={2024}}

@article{oncokb2017, author={D. Chakravarty and J. Gao and S. M. Phillips and others}, title={{OncoKB}: A precision oncology knowledge base}, journal={JCO Precis. Oncol.}, number={1}, pages={1--16}, year={2017}, note={Art. no. PO.17.00011}}

@article{civic2017, author={M. Griffith and N. C. Spies and K. Krysiak and others}, title={{CIViC} is a community knowledgebase for expert crowdsourcing the clinical interpretation of variants in cancer}, journal={Nat. Genet.}, volume={49}, number={2}, pages={170--174}, year={2017}}

@article{amp2017, author={M. M. Li and M. Datto and E. J. Duncavage and others}, title={Standards and guidelines for the interpretation and reporting of sequence variants in cancer}, journal={J. Mol. Diagn.}, volume={19}, number={1}, pages={4--23}, year={2017}}

@misc{rrsi2026, author={P. Xia and R. Han and Z. Wang and others}, title={{RRSI}: Regularized recursive self-improvement of agent harnesses}, howpublished={arXiv:2609.24972}, year={2026}}

@article{ring2022, author={A. Ring and D. Campo and T. B. Porras and others}, title={Circulating tumor cell transcriptomics as biopsy surrogates in metastatic breast cancer}, journal={Ann. Surg. Oncol.}, volume={29}, number={5}, pages={2882--2894}, year={2022}}

@article{chan2013, author={M. Chan and C. S. Liaw and S. M. Ji and others}, title={Identification of circulating {microRNA} signatures for breast cancer detection}, journal={Clin. Cancer Res.}, volume={19}, number={16}, pages={4477--4487}, year={2013}}

@article{mirge2021, author={A. H. Patil and M. K. Halushka}, title={{miRge3.0}: A comprehensive {microRNA} and {tRF} sequencing analysis pipeline}, journal={NAR Genomics Bioinform.}, volume={3}, number={3}, pages={lqab068}, year={2021}}

@article{cfdnaanalyzer2026, author={J. Zhou and K. Zhu and X. Huang and J. Yuan and Y. Li}, title={A comprehensive toolkit for analyzing cell-free {DNA} genomic sequencing data in liquid biopsy}, journal={iScience}, volume={29}, number={5}, pages={115829}, year={2026}}

@article{deseq2, author={M. I. Love and W. Huber and S. Anders}, title={Moderated estimation of fold change and dispersion for {RNA-seq} data with {DESeq2}}, journal={Genome Biol.}, volume={15}, pages={550}, year={2014}}

@article{snakemake, author={J. K{\"o}ster and S. Rahmann}, title={Snakemake---A scalable bioinformatics workflow engine}, journal={Bioinformatics}, volume={28}, number={19}, pages={2520--2522}, year={2012}}

@inproceedings{medrag2024, author={G. Xiong and Q. Jin and Z. Lu and A. Zhang}, title={Benchmarking retrieval-augmented generation for medicine}, booktitle={Findings of the Association for Computational Linguistics: ACL 2024}, year={2024}, pages={6233--6251}}

@article{coscientist2026, author={J. Gottweis and others}, title={Accelerating scientific discovery with {Co-Scientist}}, journal={Nature}, volume={655}, number={8122}, pages={487--496}, year={2026}}

@misc{dreamrsi2026, author={T. Zheng and X. Wu and Z. Zhang and others}, title={{Dream-RSI}: Recursive self-improvement through evolving worlds}, howpublished={arXiv:2609.14858}, year={2026}}

@misc{geo270497, author={{National Center for Biotechnology Information}}, title={Gene {Expression} {Omnibus}: {GSE270497}, small {RNA} sequencing of exosomes in breast cancer screening population}, howpublished={[Online]. Available: https://www.ncbi.nlm.nih.gov/geo/query/acc.cgi?acc=GSE270497}, note={Updated Jan. 30, 2025}}

@article{bh1995, author={Y. Benjamini and Y. Hochberg}, title={Controlling the false discovery rate: A practical and powerful approach to multiple testing}, journal={J. R. Stat. Soc. Ser. B}, volume={57}, number={1}, pages={289--300}, year={1995}}

@article{danos2019, author={A. M. Danos and K. Krysiak and E. K. Barnell and others}, title={Standard operating procedure for curation and clinical interpretation of variants in cancer}, journal={Genome Med.}, volume={11}, pages={76}, year={2019}}

@article{wagner2020, author={A. H. Wagner and B. Walsh and G. Mayfield and others}, title={A harmonized meta-knowledgebase of clinical interpretations of somatic genomic variants in cancer}, journal={Nat. Genet.}, volume={52}, number={4}, pages={448--457}, year={2020}}

@article{horak2022, author={P. Horak and M. Griffith and A. M. Danos and others}, title={Standards for the classification of pathogenicity of somatic variants in cancer (oncogenicity): Joint recommendations of {Clinical Genome Resource} ({ClinGen}), {Cancer Genomics Consortium} ({CGC}), and {Variant Interpretation for Cancer Consortium} ({VICC})}, journal={Genet. Med.}, volume={24}, number={5}, pages={986--998}, year={2022}}

@article{escat2018, author={J. Mateo and D. Chakravarty and R. Dienstmann and others}, title={A framework to rank genomic alterations as targets for cancer precision medicine: The {ESMO} scale for clinical actionability of molecular targets ({ESCAT})}, journal={Ann. Oncol.}, volume={29}, number={9}, pages={1895--1902}, year={2018}}

@misc{llmwiki, author={{Nashsu}}, title={llm\_wiki: Knowledge graph with relevance model}, howpublished={GitHub. [Online]. Available: https://github.com/nashsu/llm\_wiki}, note={Accessed Oct. 8, 2026}}

@article{mo2024, author={J. Mo and others}, title={A machine learning model revealed that exosome small {RNAs} may participate in the development of breast cancer through the chemokine signaling pathway}, journal={BMC Cancer}, volume={24}, pages={1435}, year={2024}}

@article{cutadapt2011, author={M. Martin}, title={Cutadapt removes adapter sequences from high-throughput sequencing reads}, journal={EMBnet.journal}, volume={17}, number={1}, pages={10--12}, year={2011}}

@article{bwa2009, author={H. Li and R. Durbin}, title={Fast and accurate short read alignment with {Burrows--Wheeler} transform}, journal={Bioinformatics}, volume={25}, number={14}, pages={1754--1760}, year={2009}}

@article{samtools2009, author={H. Li and B. Handsaker and A. Wysoker and others}, title={The {Sequence Alignment/Map} format and {SAMtools}}, journal={Bioinformatics}, volume={25}, number={16}, pages={2078--2079}, year={2009}}

@article{lofreq2012, author={A. Wilm and P. P. K. Aw and D. Bertrand and others}, title={{LoFreq}: A sequence-quality aware, ultra-sensitive variant caller for uncovering cell-population heterogeneity from high-throughput sequencing datasets}, journal={Nucleic Acids Res.}, volume={40}, number={22}, pages={11189--11201}, year={2012}}

\end{document}